\documentclass[a4paper,conference,table,xcdraw]{IEEEtran}
\usepackage[a4paper, left=37pt, right=37pt, bottom=113pt, top=54pt]{geometry}
\usepackage{multirow}
\usepackage{amsmath}
\usepackage{algorithm}
\usepackage{algpseudocode}
\usepackage{balance}
\usepackage{pifont}
\usepackage{booktabs}

\usepackage{scalerel}
\usepackage{tikz}
\usetikzlibrary{svg.path}

\definecolor{orcidlogocol}{HTML}{A6CE39}
\tikzset{
  orcidlogo/.pic={
    \fill[orcidlogocol] svg{M256,128c0,70.7-57.3,128-128,128C57.3,256,0,198.7,0,128C0,57.3,57.3,0,128,0C198.7,0,256,57.3,256,128z};
    \fill[white] svg{M86.3,186.2H70.9V79.1h15.4v48.4V186.2z}
                 svg{M108.9,79.1h41.6c39.6,0,57,28.3,57,53.6c0,27.5-21.5,53.6-56.8,53.6h-41.8V79.1z M124.3,172.4h24.5c34.9,0,42.9-26.5,42.9-39.7c0-21.5-13.7-39.7-43.7-39.7h-23.7V172.4z}
                 svg{M88.7,56.8c0,5.5-4.5,10.1-10.1,10.1c-5.6,0-10.1-4.6-10.1-10.1c0-5.6,4.5-10.1,10.1-10.1C84.2,46.7,88.7,51.3,88.7,56.8z};
  }
}

\newcommand\orcidicon[1]{\href{https://orcid.org/#1}{\mbox{\scalerel*{
\begin{tikzpicture}[yscale=-1,transform shape]
\pic{orcidlogo};
\end{tikzpicture}
}{|}}}}

\usepackage{cite}

\newif\ifquestion
\questiontrue 

\usepackage{graphicx}
\usepackage{multirow}
\usepackage{hyperref} 
\usepackage{soul}
\usepackage{amssymb}

\begin{document}
%

\title{Graph Neural Networks and Representation Embedding for Table Extraction in PDF  Documents}

\author{\IEEEauthorblockN{Andrea Gemelli \textsuperscript{\textsection} \orcidicon{0000-0002-6149-8282}}
\IEEEauthorblockA{DINFO, University of Florence\\
Florence, Italy\\
Email: andrea.gemelli@unifi.it}
\and
\IEEEauthorblockN{Emanuele Vivoli \textsuperscript{\textsection} \orcidicon{0000-0002-9971-8738}}

\IEEEauthorblockA{DINFO, University of Florence\\
Florence, Italy\\
Email: emanuele.vivoli@unifi.it}
\and
\IEEEauthorblockN{Simone Marinai \orcidicon{0000-0002-6702-2277}}

\IEEEauthorblockA{DINFO, University of Florence\\
Florence, Italy\\
Email: simone.marinai@unifi.it}}

\maketitle

\begingroup\renewcommand\thefootnote{\textsection}
\footnotetext{Andrea Gemelli and Emanuele Vivoli contributed equally.}

\thispagestyle{plain}
\pagestyle{plain}



\begin{abstract}

Tables are widely used in several types of documents since they can bring important information in a structured way. 
In scientific papers, tables can sum up novel discoveries and summarize experimental results, making the research comparable and easily understandable by scholars. 
Several methods perform table analysis working on document images, losing useful information during the conversion from the PDF files since
OCR tools  can be prone to recognition errors, in particular for text inside tables. 
The main contribution of this work is to tackle the problem of table extraction, exploiting Graph Neural Networks. Node features are enriched with suitably designed representation embeddings. These representations help to better distinguish not only tables from the other parts of the paper, but also table cells from table headers. 
We experimentally evaluated the proposed approach on a new dataset obtained by merging the information provided in the PubLayNet and PubTables-1M datasets.

\end{abstract}

\IEEEpeerreviewmaketitle


\section{Introduction}

Nowadays, scientific documents are usually shared as PDF files either containing scanned pages or born-digital contents. Since current literature is mainly available as born-digital we focus on these types of documents.
The main purpose of the PDF format is to render the document content in a faithful way on different platforms and devices. 
Despite the wide use of PDF files, 95.5\% of published articles in PDF format are not semantically tagged\ \cite{15-06-The_Portable}. Therefore, extracting information from these documents remains a difficult problem.
This is particularly important for tables that are generated taking into account the semantic information (e.g. from \LaTeX\ or MSWord) that is lost in the PDF. As a consequence, Table Extraction (TE) is still challenging; in Section\ \ref{sec:compare}, we describe TE as the task of detecting tables, recognizing their structure, and analyzing their contents (e.g. finding table headers) at once.
In academic papers, tables are commonly used as a compact and efficient way for describing statistical and relational information\ \cite{04-00-Table_structure}.
It is essential to efficiently extract and analyze them, e.g. to compare SOTA results\ \cite{20-04-AxCell}.

The first techniques for extracting information from tables explored solutions for PDF documents taking into account heuristics object-based approaches. These methods analyze both textual and positional information often relying on heuristics \cite{09-00-User-Guided, Marinai-DocEng10}.
Since 2013, competitions on table analysis have been organized to deal with both born-digital and scanned documents\ \cite{13-00-ICDAR13, 17-00-ICDAR17-POD-dataset}. 
So far, most of the works consider image-based approaches;
as shown in\ \cite{21-06-Current} recent methods often exploit Computer Vision and Natural Language Processing (NLP) techniques to deal with tables and OCR tools are employed whenever text is needed.

In this work we tackle the problem of TE in PDF scientific documents as a node classification task, exploiting Graph Neural Networks (GNNs).
This is based on recent research showing that GNNs are capable of better recognizing table layouts using structural information\  \cite{19-04-TDInvoice}.

The main contributions of the paper  \footnote{\href{https://github.com/AILab-UniFI/GNN-TableExtraction}{Code on GitHub: https://github.com/AILab-UniFI/GNN-TableExtraction}} are listed below:
     
\begin{enumerate}
    \item TE is redefined as a node classification task, addressed by a GNN.
    Graph nodes are composed of basic PDF objects while edges are computed considering relationships and mutual distances between nodes. Our experiments show that GNNs are well suited for TE;
    
    \item the graph nodes are augmented using a novel embedding of different representations for numerical and non-numerical values. These embeddings are learned over table cells elements taking into account the PubTables-1M dataset \cite{21-10-PubTables-1M}. Our experiments demonstrate the efficiency of the proposed representation embeddings in conjunction with the node positional information. Ablation studies are conducted exploring also NLP-based word embeddings;

    \item new custom annotated data are collected by merging the ground-truths of two widely-used datasets for Document Layout Analysis (DLA) \cite{19-08-PubLayNet} and for TE \cite{21-10-PubTables-1M}. This novel dataset allows us to perform both tasks at the same time. 
\end{enumerate}

The paper is organized as follows: in Section \ref{sec-related} we report some background for the method proposed. Our main contributions are described in Section \ref{sec-method}, while in Section \ref{sec-exper} we present and discuss the experiments performed. Finally, conclusions are drawn in Section \ref{sec-concl}.


\section{Related works} 
\label{sec-related}

We first analyze the main properties of tools used to extract information from PDF files. 
Then, we discuss NLP approaches for word embeddings taking into account various methods proposed to address tasks related to tables in documents. In particular, since we are using a graph-based method, we summarize the most relevant works adopting Graph Neural Networks on documents. Lastly, we clarify the task of Table Extraction addressed by our work.

\subsection{Information extraction from PDF files}
\label{ref:SOA-PDF}

Since scientific papers are currently distributed as born-digital PDF files it is appropriate, in our view, to look at the information in the PDF file without relying on error-prone OCR tools\ \cite{22-02-OCR} \cite{21-11-DUE}.
However, extracting information from PDF documents is not easy and various PDF parsers with different features can be used for this purpose.

A PDF file is a combination of basic objects, including text, vector graphics, and images. 
Typographic features of these objects allow visualization tools to render the document consistently on different devices.  Unfortunately, the typographic information hinders the extraction of the document content since the semantic of objects is often lost in the PDF.

Many high-level tools for PDF object extraction are available. Some only extract basic PDF elements such as text, images, and graphical items \cite{12-00-PyMuPDF}. Others are able to extract additional information specifically for scientific papers, such as title, authors, and abstract \cite{05-00-GROBID} \cite{20-00-S2orc-doc2json} \cite{Marinai-ICDAR09}. In other cases, it is possible to extract more complex items (e.g. tables) \cite{16-06-Camelot}, but these tools can require an accurate location of the table or they can fail when tables are surrounded by text \cite{15-05-tabula-java} \cite{20-11-PDFScraper}.
The latter techniques often rely on low-level tools for PDF parsing (e.g. PDFMiner \cite{11-07-PDFMiner.six} MuPDF \cite{04-09-MuPDF}, or PDFBox \cite{08-03-Apache.PDFBox}).

In our work, we use PyMuPDF\ \cite{12-00-PyMuPDF} because it is a reliable and well-documented tool that provides the token-level objects we need to build the graph.

\subsection{Text and numeral embeddings}
\label{sec:SOA-embed}

Even though tables in documents can be recognized by only using the layout information of document objects, textual information can help for this task. 
Common methods for representing textual information rely on word embeddings. 
Static embeddings represent isolated words (i.e. Word2Vec \cite{13-01-Word2Vec}, GloVe \cite{14-10-GloVe}) while contextualized embeddings provide different word representations according to different contexts (i.e. Elmo \cite{18-02-Elmo}, BERT \cite{18-10-BERT}). 
Word embeddings are extremely useful in several NLP tasks; however, it is difficult to represent numbers, formulas, and intervals that often appear in tables. 
Recently, \cite{20-11-Learning} defined a new Word2Vec approach for enhancing numerical embeddings.
Word embeddings can hardly learn numerals as there are an infinite number of them and their individual appearances in training corpora are rare. 
Two different representations are provided for words and numerals: the latter are represented by prototypes obtained by clustering numerals with SOM or GMM. Numerals are represented either by the closest prototype or by a weighted average of the closest prototypes. 

Inspired by \cite{20-11-Learning} we propose a representation embedding to handle formulas and intervals in addition to words and numerals as described in Section\ \ref{sec-repr}.

\subsection{Methods and tasks related to tables}
\label{ref:SOA-tables}

Table Understanding (TU) consists of three steps\ \cite{21-06-Current}: Table Detection (TD), Table Structural Segmentation (TSS), and Table Recognition (TR).  
TSS is referred to as Table Structure Recognition (TSR) in\ \cite{21-10-PubTables-1M} where the recognition of column and projected row headers is defined as Table Functional Analysis (TFA).
To perform TD table boundary coordinates are detected. 
This task is often performed in the image domain\ \cite{Cesarini-ICPR02} and recently approached with object detection techniques. Usually, models like Faster-RCNN and Mask-RCNN \cite{19-08-PubLayNet} \cite{19-08-TableBank} are used. On the other hand, at the token level, NLP-based methods are involved in using both textual and visual features, such as LayoutLM \cite{19-12-LayoutLM} in \cite{20-06-DocBank}.
Once tables are found, their structure is recognized by identifying their
rows, columns, and cell positions. 
A Cascade Mask R-CNN (CascadeTabNet) is used in\ \cite{20-04-CascadeTabNet} to detect tables and body cells, arranging them in columns and rows based on their positions.  

However, these methods do not take into account the document structure.
Graph Neural Networks (GNNs)\ \cite{09-03-GNNModel}\ \cite{17-07-GDL} have been used in several domains including DLA and TU, since they can suitably represent the layout structures and object positions that otherwise would be lost by the aforementioned models.

TD in invoices by using GNNs has been proposed firstly in \cite{19-04-Table_understanding}. Graph nodes correspond to word boxes and are enriched with geometrical, textual, and image features; edges connect nodes following the reading order.
TSR in segmented tables is addressed in \cite{19-07-Rethinking} by classifying edges for cells, rows, and columns.
Each node uses only positional features, but the authors suggest that using also NLP features would result in a method improvement.
Edge information is considered in\  \cite{19-04-TDInvoice} to enrich the structural information in the computation of node features.
More recently, \cite{21-06-TGRNet} reformulate the problem of TSR as an end-to-end table reconstruction through node classification. For each node, spatial and logical information (start-end rows and start-end columns) are considered.
In the ICDAR 2021 competition on DLA and TR \cite{21-09-ICDAR}, the DAVAR-Lab-OCR obtained SOTA results on both tasks by using two different approaches: VSR\ \cite{21-09-VSR} and LGPMA\ \cite{21-09-LGPMA}.
In VSR, visual, semantic, and structural features are combined to detect objects, exploiting a GNN for the final refinement; LGPMA uses a soft pyramid mask learning mechanism in both local and global feature maps to recover the table structure, also taking into account empty cells location. Even if the structure is considered, tasks regarding tables are still carried out separately.
 
As seen so far, graphs can bring advantages beyond solely relying on visual and/or language features. 
Moreover, none of the operations shown dealing with tables are carried out at once but are processed in separated subsequent steps. 
In Section \ref{sec-method} we present our approach and we show how it can handle DLA, TD, and TFA at once. We use an approach at the token level, exploiting GNNs and taking advantage of structural information and representation embeddings.

\begin{table}
\centering
\caption{Comparison of tasks performed by different methods.\\ 
(*) future extension of the proposed method.}
\label{tab-compare}
\vspace{-0.2cm}
\resizebox{\linewidth}{!}{%
\begin{tabular}{l|cccc}
\multicolumn{1}{c}{\multirow{2}{*}{\tiny{Methods}}} & \multicolumn{4}{c}{\tiny{Tasks}} \\
\multicolumn{1}{l}{} & \tiny{DLA} & \tiny{TD} & \tiny{TSR} & \tiny{TFA} \\ 
\hline
\vspace{-0.08cm} 
\tiny{LayoutLM} \cite{20-06-DocBank} & \tiny{\checkmark} & \tiny{\checkmark} &  &  \\
\vspace{-0.08cm}
\tiny{VSR} \cite{21-09-VSR} & \tiny{\checkmark} & \tiny{\checkmark} &  &  \\
\vspace{-0.08cm}
 \tiny{Riba et al.}\cite{19-04-TDInvoice} &  & \tiny{\checkmark} &  &  \\
\vspace{-0.08cm}
 \tiny{CascadeTabNet}\cite{20-04-CascadeTabNet} &  & \tiny{\checkmark} & \tiny{\checkmark} &  \\
\vspace{-0.08cm}
 \tiny{LGPMA}\cite{21-09-LGPMA} &  &  & \tiny{\checkmark} &  \\
\vspace{-0.08cm}
 \tiny{TGRNet}\cite{21-06-TGRNet} &  &  & \tiny{\checkmark} &  \\ 
\hline
\tiny{Ours} & \tiny{\checkmark} & \tiny{\checkmark} & {\tiny{(\checkmark*)}} & \tiny{\checkmark} \\ 
\hline
\end{tabular}
}
\end{table}

\subsection{Problem Formulation}
\label{sec:compare}

The term Table Extraction (TE) appears back in 2003 \cite{03-00-Table-Extraction}, aiming at labeling each line of a document with a tag and describing its function relative to tables. 
A more recent work \cite{21-10-PubTables-1M} proposes another meaning for TE, providing TD, TSR, and TFA annotations.
With TE, we refer to the task of detecting tables and extracting the meaning of their content at once, at the token level.
To this purpose, we adopt a GNN to tackle TE as a node classification problem. 
To the best of our knowledge, this method is the only one addressing table extraction and document layout analysis at once. In Table \ref{tab-compare} we compare the sub-tasks performed by some methods  described in Section\ \ref{ref:SOA-tables}.  The proposed framework can address also TSR, but implementation details are still under investigation.

\begin{figure*}[t]
    \centering
    \includegraphics[width=\textwidth]{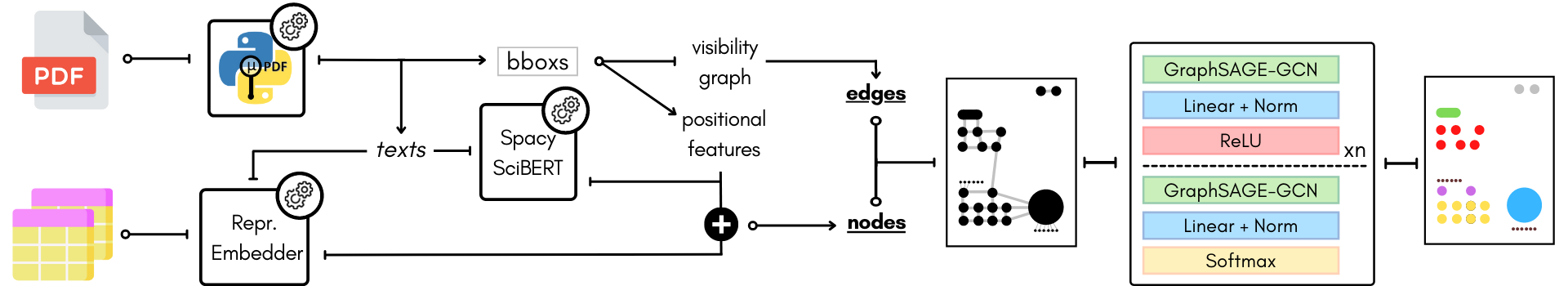}
    \caption{Overview of the proposed method.}
    \label{fig:network}
\end{figure*}

\section{Proposed Method}
\label{sec-method}

In this Section, we describe the main components of the proposed method
(Fig. \ref{fig:network}). 
First, we present the dataset building, both considering the data collection and the generation of annotations. 
Then, we illustrate how a PDF paper is handled to build its graph, adding explanations about node and edge features and representation embeddings.
At the end we describe the message passing algorithm applied to train the GNN and the strategies used to handle class imbalance during training.

\subsection{Dataset generation}
\label{sec-data-gen}
Although it is easy to freely access large collections of scientific papers (i.e. from arXiv or PubMed Central), it is difficult to find documents labeled with complete information.
Most benchmark datasets support either document layout analysis or table understanding, however, our aim is to perform the first both tasks in a unified way, and therefore we need a new dataset.
To do so, we merged the data and the annotations given by the PubLayNet and PubTables-1M datasets, both based on PubMed Central publications. 
PubLayNet is a collection of $358,353$ PDF pages with five types of regions annotated (\textit{title, text, list, table, image})  \cite{19-08-PubLayNet}. 
PubTables-1M \cite{21-10-PubTables-1M} is a collection of $947,642$ fully annotated tables, including information for table detection, recognition and functional analysis (such as \textit{column headers, projected rows} and \textit{table cells}). The datasets are built to address different tasks as summarized in Table \ref{tab:dataset}. 

\begin{table}
\centering
\caption{Comparison of original and merged datasets (Document Layout Analysis (DLA), Table Detection (TD), Table Structure Recognition (TSR), and Table Functional Analysis (TFA)).}
\resizebox{\linewidth}{!}{%
\begin{tabular}{l|cccc}
Datasets & \begin{tabular}[c]{@{}c@{}}\# pages\\~(train / val / test)\end{tabular} & \# tables & tasks & \# classes \\ \hline
& & & & \\
\vspace{-0.1cm} PubLayNet & {336k /~}{11k / 11k} & 107k & DLA, TD & 5 \\
& & & & \\
\vspace{-0.1cm}  PubTables-1M &  {460k / 57k / 57k} & 948k & TD, TSR, TFA & 7 \\
& & & & \\\hline \hline
& & & & \\
\vspace{-0.1cm}  Merged & 67k / 1.5k / 1.5k & 27k & DLA, TD, TSR, TFA & 13 \\
& & & & \\\hline
\end{tabular}
}
\label{tab:dataset}
\end{table}

To merge the datasets,  we first identify papers belonging to both collections.
Then, from this subset, we keep pages with tables fully annotated in PubTables-1M and pages without tables.
The numbers of pages in the datasets are shown in Table \ref{tab:dataset}. 
The merged dataset contains 13 different classes adding to the regions annotated in PubLayNet, the table annotations described in PubTables-1M (\textit{row, column, table header, projected header, table cell}, and \textit{grid cell}).
Moreover, we add two classes: \textit{caption} and \textit{other}. \textit{Captions} are heuristically found taking into account the proximity with images and tables, while the \textit{other} class contains all the remaining not-labeled text regions (e.g. page headers and page numbers).

\begin{figure}[b]
    \centering 
    \includegraphics[width=.45\textwidth]{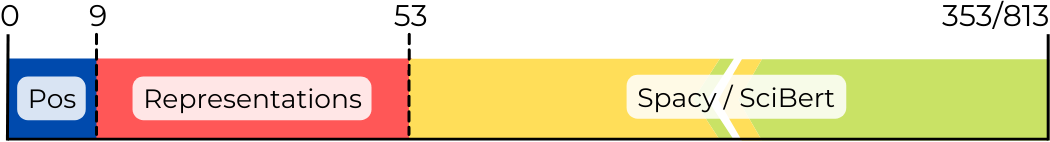}
    \caption{Node features: positions and representations have a fixed length;  Spacy or SciBERT embeddings have a length of 300 and 760, respectively.}
    \label{fig:fv}
\end{figure}

\subsection{Converting PDF pages to graphs}
\label{sec-graph} 
Graphs are generated from PDF files in three steps:
\begin{enumerate}
    \item information about basic items (tokens) in PDFs are extracted by using PyMuPDF \cite{12-00-PyMuPDF}. 
    \item each node is connected to its nearest visible nodes according to the visibility graph\ \cite{19-04-TDInvoice};
    \item features are added to each node and edge. 
\end{enumerate}

Following ideas presented in Section\ \ref{ref:SOA-tables} that have been proven successful for TD \cite{19-04-Table_understanding} and TSR \cite{ 19-07-Rethinking} we enrich our graph nodes with positional and textual features. We use new representation embedding features (Section\ \ref{sec-repr}) that help the model to better discriminate table cells and headers from the rest (performing TFA). Inspired by\ \cite{19-04-TDInvoice} we make use of edge weights and enrich graph edges with token boxes distances, letting closer elements contribute more to the message passing algorithm (Section \ref{sec-mess-pass}).

Node features are a combination of geometrical and textual  information as shown in Fig. \ref{fig:fv}.
The geometrical features are  $<x_1, y_1, x_2, y_2, w, h, x_c, y_c, A>$, 
where $x_1, y_1, x_2, y_2$ are the corners of the bounding box having width $w$, height $h$, center $x_c, y_c$ and area $A$.
Other node features describe the textual content from different perspectives: 
a) inspired by \cite{19-04-TDInvoice} we add
three values $<$\% of characters, \% of digits, \% of  symbols$>$ to better distinguish items in tables from other page contents. The values are the percentage of characters, digits, and symbols in the node, respectively; 
b) we add a boolean value for $images$ that identifies images recognized by PyMuPDF;
c) tokens are described with the proposed representation embedding: as described in Section\ \ref{sec-repr} they are more informative if containing digits or symbols; 
d) static NLP-based embeddings (SciBERT \cite{19-03-SciBERT} and Spacy \cite{16-00-Spacy}) are also considered among node features.

\begin{algorithm}[t]
\caption{Edge weight $w_e$}\label{alg:weight}
\begin{algorithmic}
\Require $u,v \in V, e = (u,v) \in E$
\If{$(u$ above $v) \vee (u$ below $v)$}
    \State $d_e \gets \max(u.y_1, v.y_1) - \min(u.y_2, v.y_2)$
\ElsIf{$(u$ left of $v) \vee (u$ right of $v)$}
    \State $d_e \gets \max(u.x_1, v.x_1) - \min(u.x_2, v.x_2)$
\EndIf

$w_{e} = 1 - \frac{d_{e}}{\max_{e \in E}\{d_e\}}$
\end{algorithmic}
\label{algo-edge}
\end{algorithm}

The feature of edge $(u,v)$ is the distance between bounding boxes of $u$ and $v$ as defined in Algorithm\ \ref{alg:weight}.
To handle the graph, we use the DGL open-source library \cite{20-00-DGL}. 
An example of the graph corresponding to a portion of a page is shown in Fig. \ref{fig:pdf2graph}.

\begin{figure}[b]
    \centering
    \fbox{
\includegraphics[width=.36\textwidth]{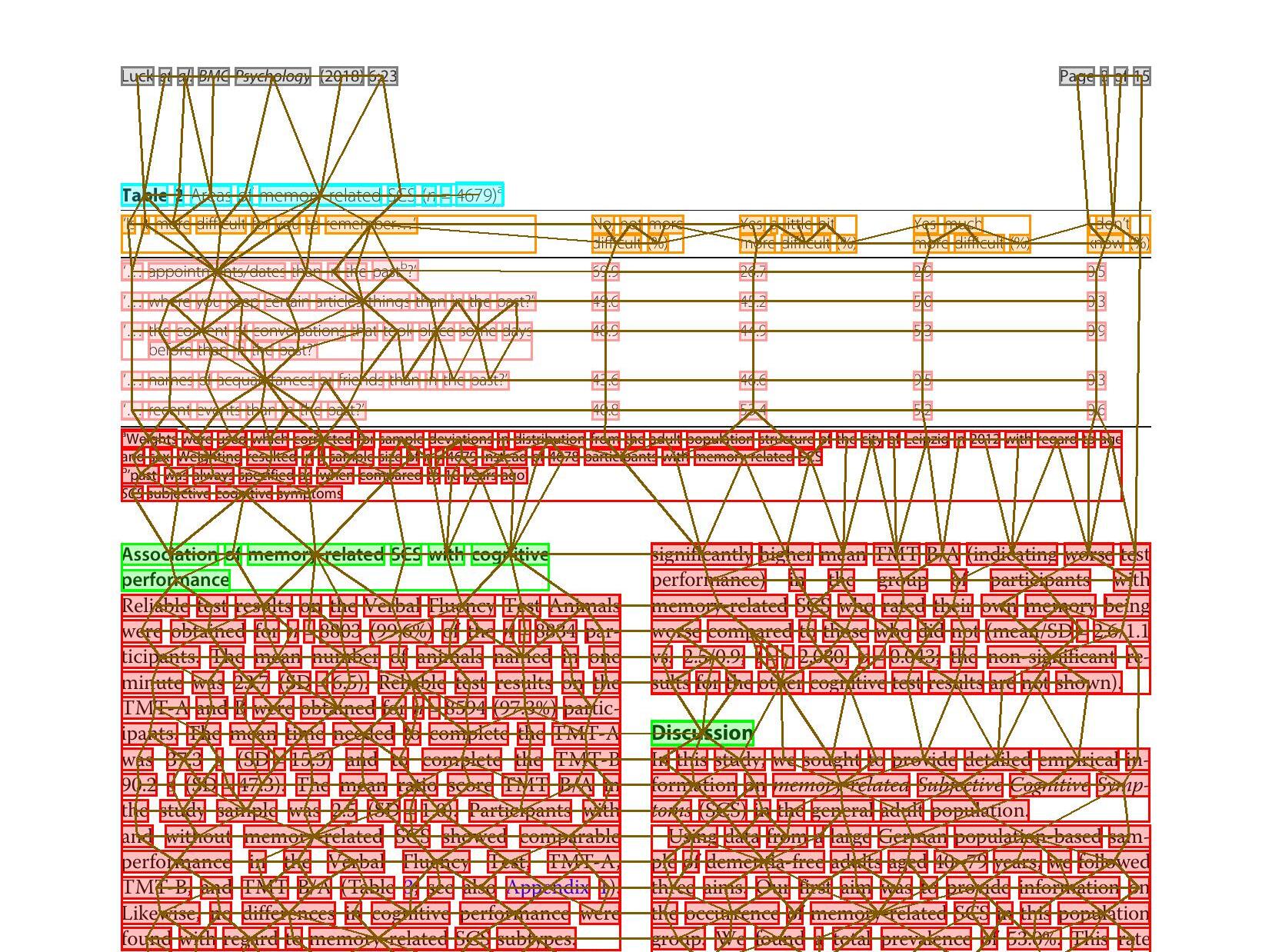} 
    }
    \caption{Graph of a portion of page. Different types of nodes have different colors. (red: text, green: title, pink: table cell, orange: table header, light blue: caption, grey: other)}
    \label{fig:pdf2graph}
\end{figure}

\subsection{Representation embedding}
\label{sec-repr}

Table headers are often either words or word-numeral combinations, while table cells mainly contain numerals or a combination of numerals with other symbols (e.g. numbers and intervals). In some cases (e.g. in tables comparing SOTA papers)  all the cells contain words, but this is not very frequent in our dataset.

For each token corresponding to a graph node, we obtain the representation embedding by first mapping the token into a standardized representation and then embedding the representation in a dense vector.
A representation is a combination of symbols (e.g. '$\pm$', '+', '$^{\circ}$') and the $x$ and $w$  characters that correspond to sequences of digits and words, respectively.
Examples of representations of tokens are: $"${\em Precision-Recall}$" \rightarrow "w$-$w"$;  $"12.5" \rightarrow "x.x"$, "+3.1(2.5$\pm$ 1.0)"$ \rightarrow "+x.x(x.x\pm x.x)"$.
Algorithm \ref{alg:word2repr} describes the function \textit{word2repr} that maps a token ($word_i$) to its representation ($repr_i$).

\begin{algorithm} [t]
\caption{Word to Representation}\label{alg:word2repr}
\begin{algorithmic}
\Require $len(word_i) > 0$
\Ensure $repr_i = word2repr(word_i)$
\State $repr_i \gets word_i$
\State $repr_i \gets repr_i.replace(/[A-Za-z]/g,\text{"w"})$
\State $repr_i \gets repr_i.replace(/[0-9]/g,\text{"x"})$
\State $repr_i \gets repr_i.sub(r\text{"(.)\textbackslash1+"}, r\text{"\textbackslash1"})$
\end{algorithmic}
\end{algorithm}

In PubTables-1M there are more than 50,000 different representations that provide an overview of the various contents of a scientific table. 
To embed the representations, we induce a set $P$ of prototypes obtained by clustering the $l = 2000$ most frequent representations in the dataset.
The clustering is obtained by computing with the Levenshtein distance \cite{89-00-String} the distance matrix $D_{l \times l}$ (corresponding to distances between representations).   
The Affinity propagation algorithm is then applied to the  $D_{l \times l}$ matrix to compute the $P$ prototypes (in our experiments we have $P=47$).


Following \cite{20-11-Learning}, we define a similar process to embed token representations by training a Word2Vect model with SkipGram negative sampling.
To train the Word2Vect over the training set $T$ of tables, we consider three ways to \textit{visit} table cells and define context elements and the target one.
Given a table $t \in T$, and considering a sliding window of size $w = 5$,  we extract for each cell $c_{i,j}^{t}$ ($i$= row, $j$= column of table $t$) a list of neighboring cells  arranged following one of three patterns:\\
\ \\
{\bf Headers}: $c_{i,j}^{t} \Rightarrow [c_{i,0}^{t}, c_{i,j-1}^{t}, c_{i,j}^{t}, c_{i-1,j}^{t}, c_{0,j}^{t}]$\\
{\bf Rhombus}: $c_{i,j}^{t} \Rightarrow [c_{i,j-1}^{t}, c_{i-1,j}^{t}, c_{i,j}^{t}, c_{i+1,j}^{t}, c_{i+1,j+1}^{t}]$\\
{\bf Linear} : $c_{i,j}^{t} \Rightarrow [c_{i,j-2}^{t}, c_{i,j-1}^{t}, c_{i,j}^{t}, c_{i,j+1}^{t}, c_{i,j+2}^{t}]$
\ \\

For instance, for the table in Fig. \ref{fig:example}, taking $c_{3,2}^t = v_9$ we obtain the following patterns:

\ \\
{\bf Headers}: $c_{3,2}^{t}= v_9 \Rightarrow [r_c, v_8, v_9, v_{5}, h_b]$\\
{\bf Rhombus}: $c_{3,2}^{t}= v_9 \Rightarrow [v_8, v_5, v_9, v_{13}, v_{10}]$\\
{\bf Linear} : $c_{3,2}^{t}= v_9 \Rightarrow [r_c, v_8, v_9, v_{10}, v_{11}]$
\ \\

In the current system we use the Rhombus method since it provides better results and most likely reflects the graph structure for a center node. 
Once we have the embeddings for the $P$ prototypes, the ones for other representations can be computed by associating each of them to its closest prototype embedding or by computing a weighted average of all the prototypes. 
In some preliminary tests, we found that the first approach provides more informative vectors. 
The resulting representation embeddings are employed in the node feature vectors.

\begin{figure}[t]
    \centering
    \fbox{
    \begin{tabular}{c|cccc}
    &  $h_a$ & $h_b$ & $h_c$&  $h_d$ \\\hline
     $r_a$ & $v_0$ & $v_1$ & $v_2$ & $v_3$   \\
    $r_b$  & $v_4$ & $v_5$&  $v_6$&  $v_7$   \\
    $r_c$  & $v_8$ & $v_9$  & $v_{10}$& $v_{11}$  \\
    $r_d$ & $v_{12}$ & $v_{13}$ &$v_{14}$ &$v_{15}$  \\
    \end{tabular}}
    \caption{Example of table to illustrate representation embedding.}
    \label{fig:example}
\end{figure}

\subsection{Message passing}
\label{sec-mess-pass}

We use GNNs because tables are naturally structured items in documents and GNNs can take advantage of structural information. 
In this work we apply a variant of the GraphSAGE algorithm \cite{17-12-SAGE}, an inductive extension of Graph Convolutional Network (GCN) proposed in \cite{17-04-GCN}, called GraphSAGE-GCN. 
The information flows through the graph aggregating node features from neighbors. 
As this process iterates, nodes incrementally gain more information from farther ones.
Given a graph $G = (V, E)$ each node $v \in V$ has its own feature vector $h_v$ (Figure \ref{fig:fv}) and it collects information from  neighbors $N(v)$, whose vectors are called {\em messages}.
Most algorithms copy the {\em messages} into the so-called {\em mailbox}, but in our case we scale them by an edge weight $w_e$ computed in Algorithm\ \ref{algo-edge}. 
The weight $w_e$ is the normalized spatial distance of nodes $u$ and $v$, it reaches the maximum value ($w_e = 1$) on touching nodes. In so doing, nearest nodes contribute more to the information flow under the hypothesis that local nodes often belong to the same class.
At step $k$, each feature vector in the neighborhood of $v$ is collected in its mailbox $m_v^k = \{w_eh_u^{k-1} \ | \ \forall u \in N(v)\}$.
Then each node aggregates messages using a permutation-invariant differentiable function, such as pooling, mean, or sum. 
We sum messages to compute a weighted average of feature vectors of neighboring nodes: 
$h_{N(v)}^k = \frac{\sum\limits_{m \in m_v^k} m}{|N(v)|} $
We then  concatenate the current node feature vector $h_v^{k-1}$ with the aggregation of neighboring nodes $h_{N(v)}^k$  and each node updates itself: $h_v^k = \sigma ( W \times \textit{CONCAT}(h_v^{k-1}, h_{N(v)}^k))$ where $W$ is the weights matrix of a fully connected layer, applied to learn different patterns in feature vectors.
Other approaches add $h_v^{k-1}$ directly to messages or just update it with $h_{N(v)}^k$.

\begin{figure}[t]
    \centering
    \fbox{
     \includegraphics[width=.3\textwidth]{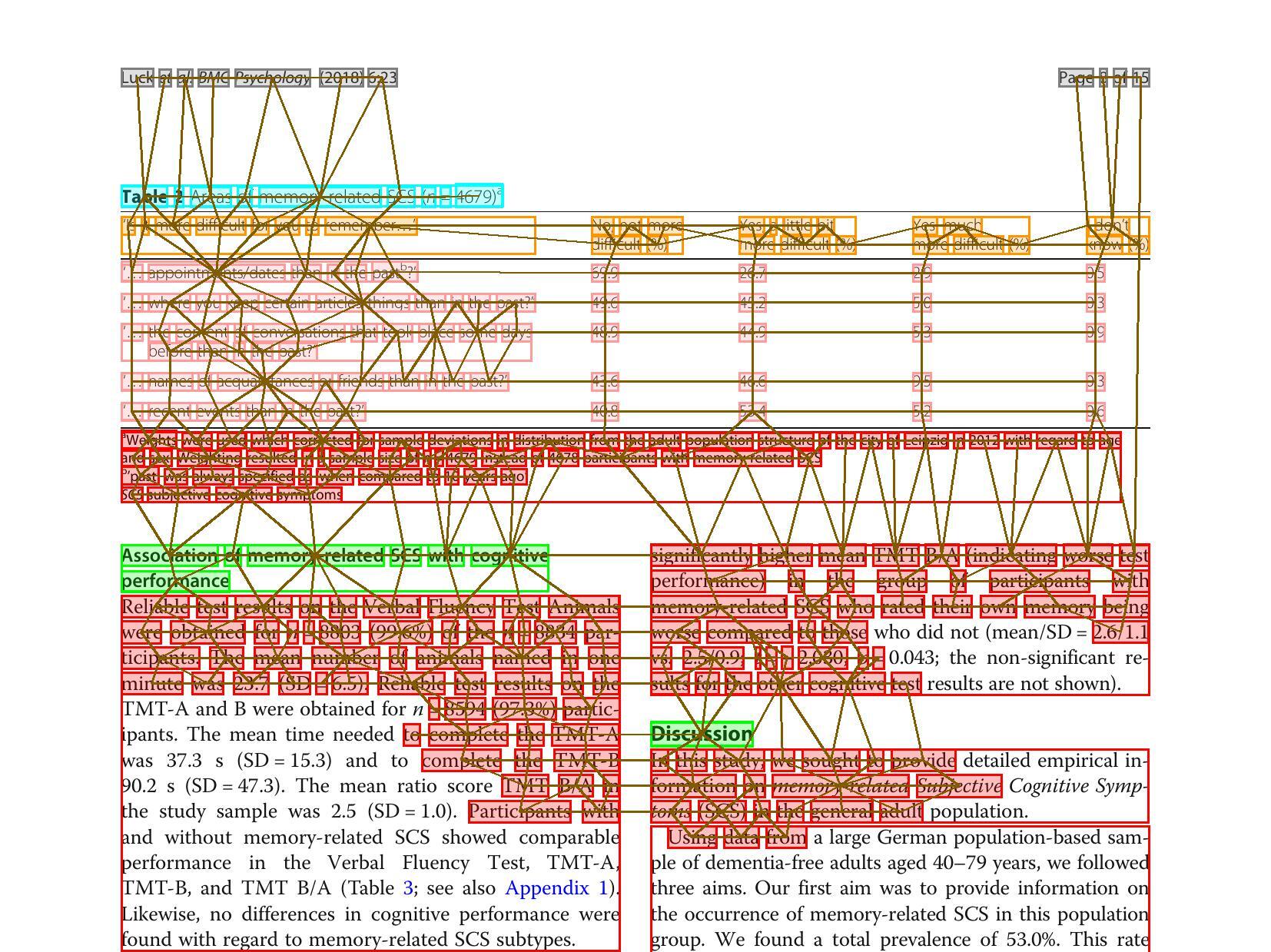}
    }
    \caption{Graph of a portion of page with text nodes in islands removed.}
    \label{fig:islands}
\end{figure}

\subsection{Class balancing by graph cutting}
\label{sec-islands}

In scientific papers, most of the nodes belong to paragraphs and are labeled as ``text'' and correspond to more than 80\% of the whole dataset. 
During training, we deal with this class imbalance by excluding pages without tables and by discarding some ``text'' nodes in the remaining pages.
A ``text'' node $v$ is discarded if there is a path with more than $k$ edges from $v$ to any other node $u$ with a different label in the original graph. Discarded nodes are called "islands".
By removing islands it is possible to reduce the number of nodes 
surrounded by others of the same class. 
In this way, the message passing algorithm aggregates more messages coming from different sources helping the method to discriminate objects.
In Fig. \ref{fig:islands} we show the graph obtained by removing islands in the graph of Fig. \ref{fig:pdf2graph}.

\begin{figure*}[t]
    \centering
    \fbox{
    \begin{tabular}{c|c|c|c}
     \includegraphics[width=.22\textwidth]{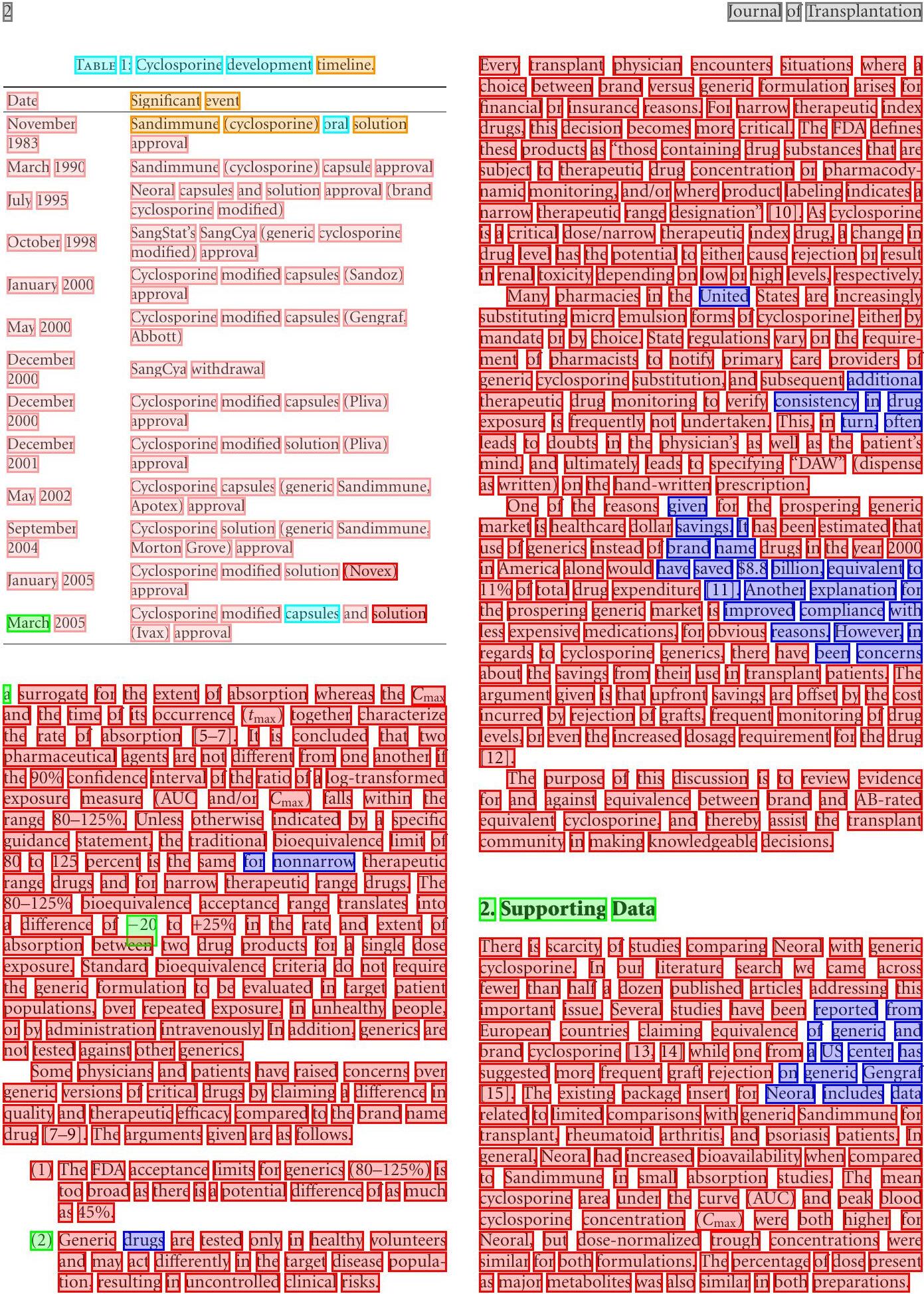} &
     \includegraphics[width=.22\textwidth]{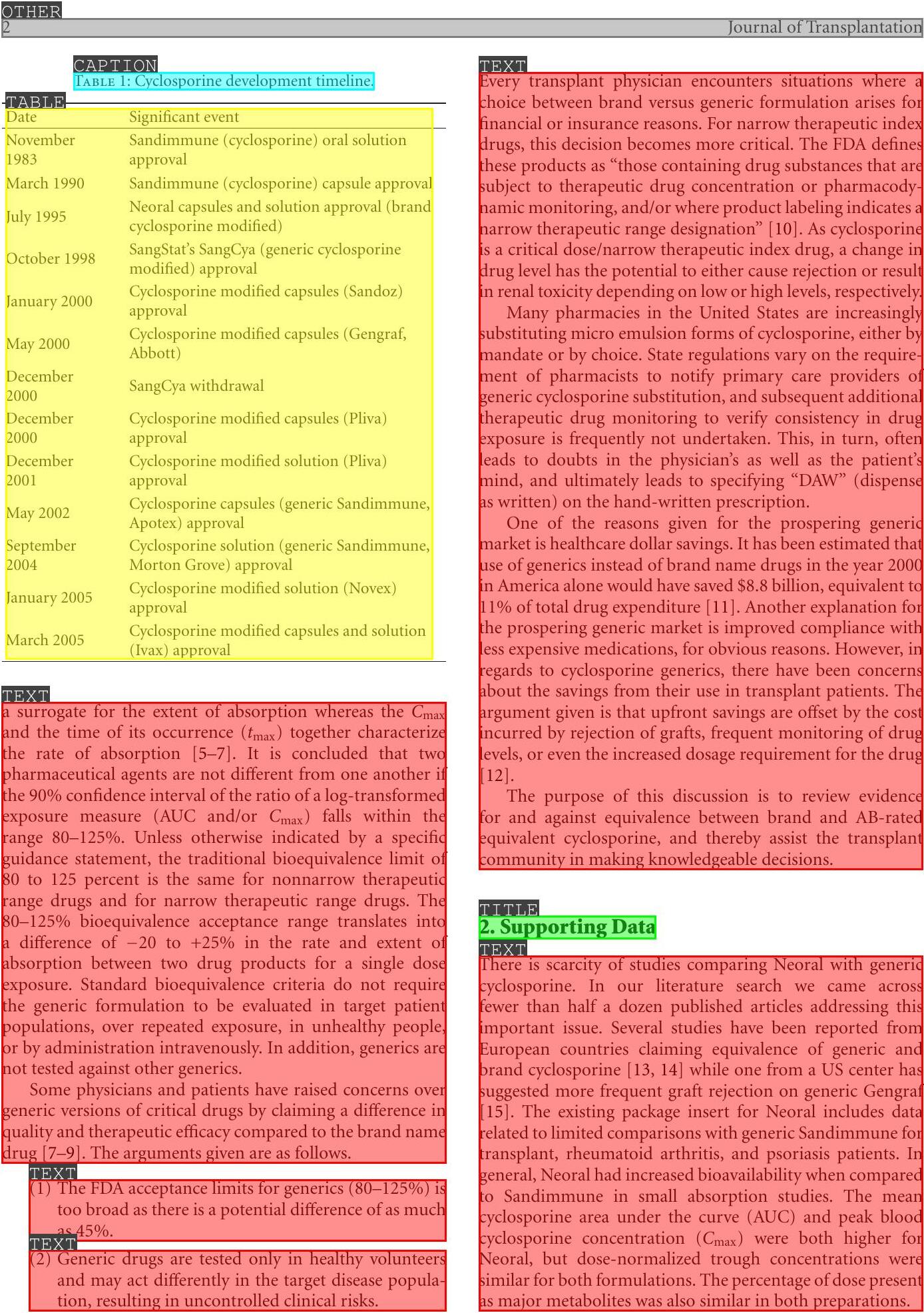} &
     \includegraphics[width=.22\textwidth]{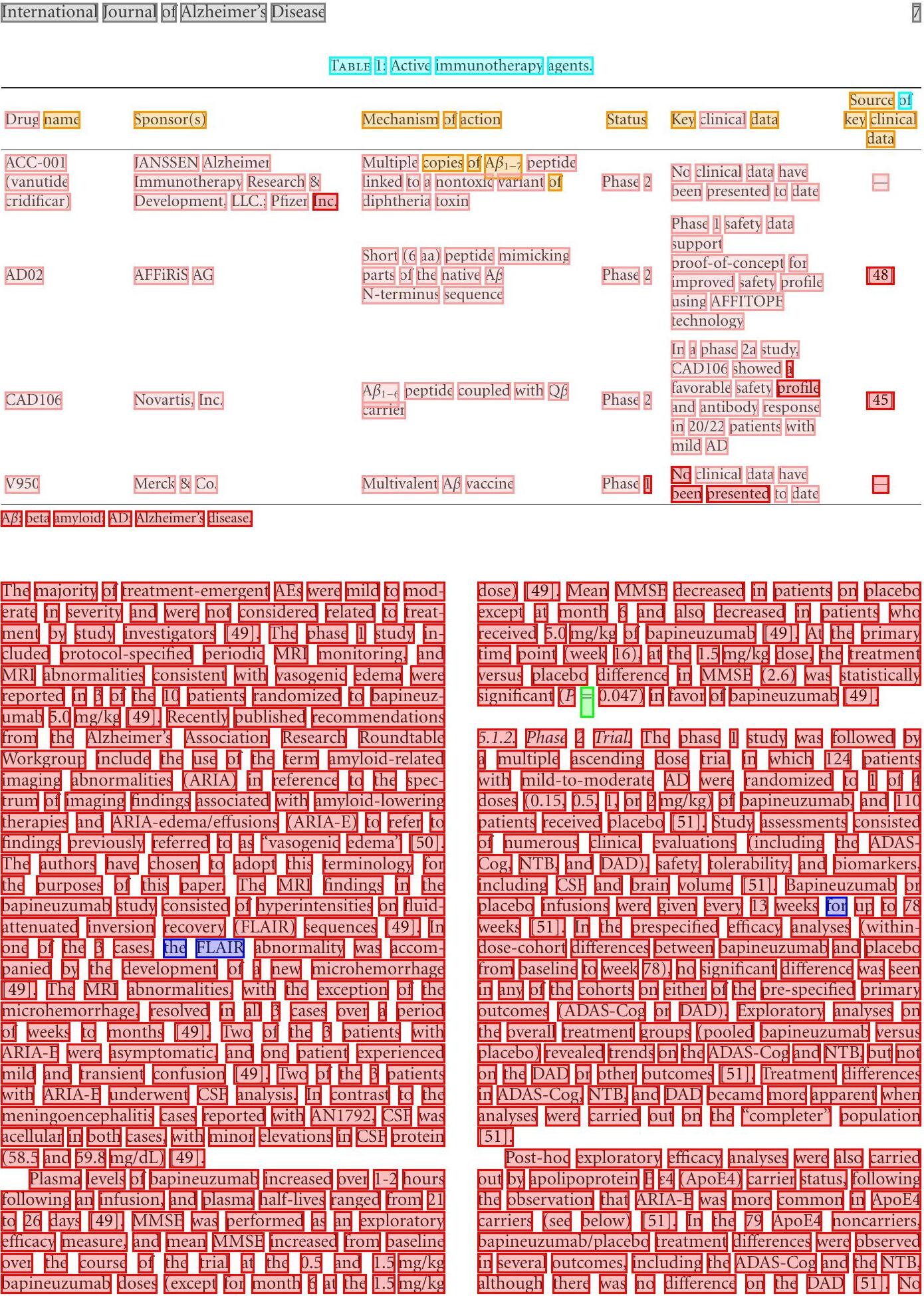} &
     \includegraphics[width=.22\textwidth]{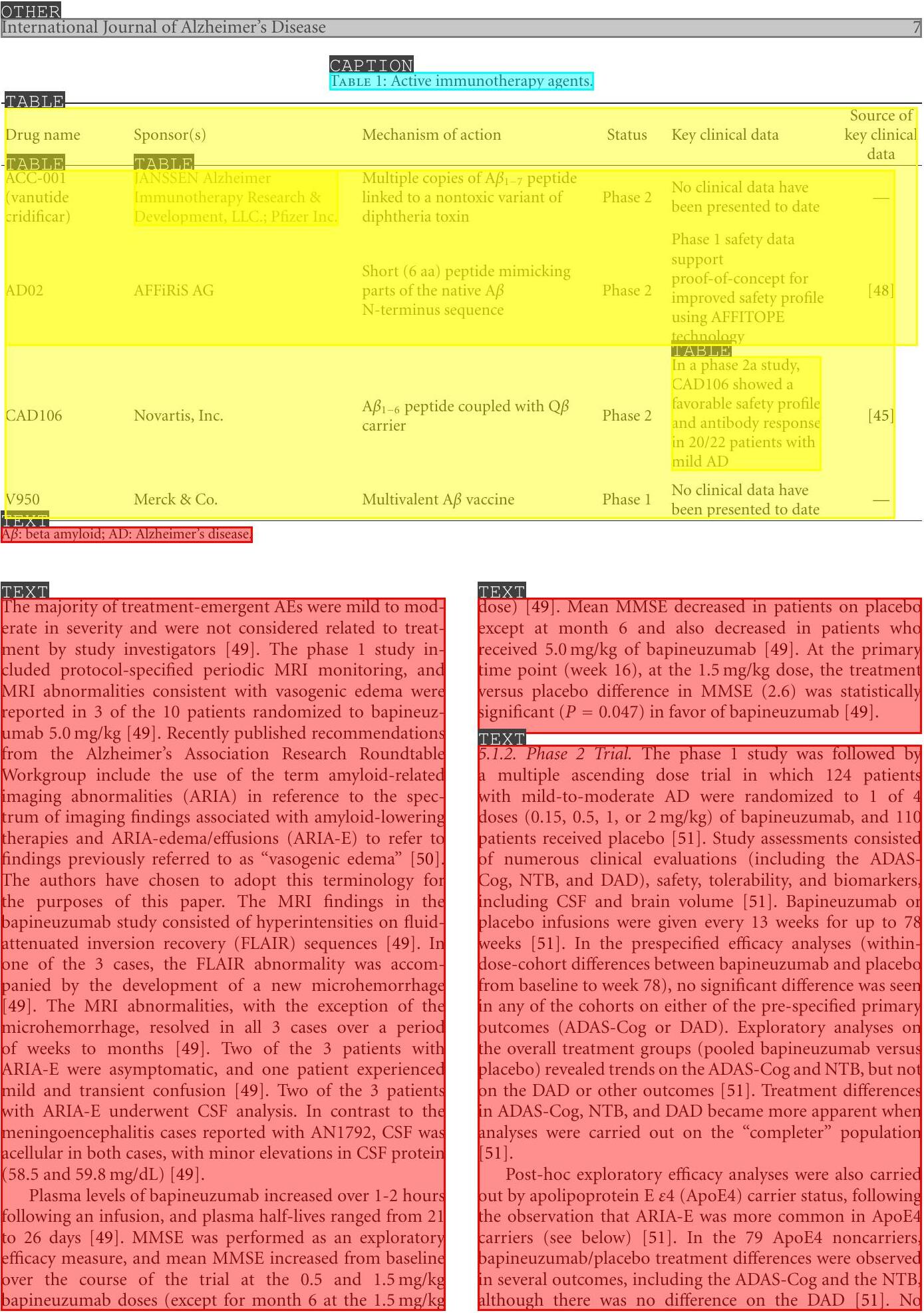}
    \end{tabular}
    }
    \caption{Examples of model inference at token-level (first and third) and post-processing (second and fourth) on two test pages. Different colors represent different types of nodes or blocks (red: text, green: title, pink: table cell, orange: table header, light blue: caption, grey: other, dark blue: list, yellow: table).}
    \label{fig:inference}
\end{figure*}



\begin{table}[b]
\caption{Evaluations are conducted with accuracy on all classes and F1 score on table cells and headers. Each model (A, B, ..) has been tested with different combination of features.}
\label{tab:experiment}
\centering
\begin{tabular}{l|l|l|lll}
\multirow{2}{*}{Model} &
\multirow{2}{*}{Features} &
  \multirow{2}{*}{Metrics} &
  \multicolumn{3}{c}{Methods} \\ \cline{4-6} 
 & &
   &
  \multicolumn{1}{l|}{\em Base} &
  \multicolumn{1}{l|}{\em Padding} &
  \multicolumn{1}{l}{\em Scaled} \\ \hline \hline
\multirow{3}{*}{$A$} & \multirow{3}{*}{bbox}      & accuracy  & 0.873 & 0.841 & 0.866 \\
                          & & cell F1   & 0.798 & 0.765 & 0.799 \\
                          & & cell-h F1 & 0.659 & 0.651 & 0.642 \\ \hline
\multirow{3}{*}{$B$}    &\multirow{3}{*}{\begin{tabular}[c]{@{}l@{}}bbox\\ + repr\end{tabular}}    & accuracy  & 0.876 & \textbf{0.875} & 0.873 \\
                          & & cell F1   & 0.821 & 0.819 & 0.816 \\
                          & & cell-h F1 & 0.653 & 0.649 & 0.648 \\ \hline \hline  
                          
\multirow{3}{*}{$C$}   &\multirow{3}{*}{\begin{tabular}[c]{@{}l@{}}bbox\\ + Spacy\end{tabular}}   & accuracy  & 0.859 & 0.847 & 0.868 \\
                          & & cell F1   & 0.767 & 0.773 & 0.781 \\
                          & & cell-h F1 & 0.685 & 0.675 & 0.660 \\ \hline
\multirow{3}{*}{$D$} &\multirow{3}{*}{\begin{tabular}[c]{@{}l@{}}bbox\\ + repr\\ + Spacy\end{tabular}}  & accuracy  & 0.865 & 0.860 & 0.809 \\
                          & & cell F1   & 0.780 & 0.776 & 0.811 \\
                          & & cell-h F1 & \textbf{0.689} & 0.675 & 0.644 \\ \hline \hline                       
\multirow{3}{*}{$E$} &\multirow{3}{*}{\begin{tabular}[c]{@{}l@{}}bbox\\ + SciBERT\end{tabular}} & accuracy  & \cellcolor[HTML]{C0C0C0}\textbf{0.882} & 0.843 & \textbf{0.879} \\
                          & & cell F1   & 0.838 & 0.816 & \textbf{0.846} \\
                          & & cell-h F1 & 0.688 & \cellcolor[HTML]{C0C0C0}\textbf{0.699} & \textbf{0.686} \\ \hline

\multirow{3}{*}{$F$} &\multirow{3}{*}{\begin{tabular}[c]{@{}l@{}}bbox\\ + repr\\ + SciBERT\end{tabular}}  & accuracy  & 0.709 & 0.787 & 0.870 \\
                          & & cell F1   & \cellcolor[HTML]{C0C0C0}\textbf{0.855} &  \textbf{0.832} & 0.777 \\
                          & & cell-h F1 & 0.668 & 0.636 & 0.671  
\end{tabular}
\end{table}

\section{Experiments}
\label{sec-exper}

In this section, we discuss the performed experiments.
At training time, only pages containing tables are used (around 27k) reserving 5\% of them for validation. The test set contains 1.5k pages including also pages without tables.
To evaluate the performance of the proposed method, model accuracy and F1 scores are considered. The F1 metric is used for table cells and table header classes. 
In this section, we discuss the methods and the corresponding results reported in Table \ref{tab:experiment}.

\subsection{Methods}
\label{sec-ours}
Each method differs from the others varying the hidden layer dimensions $h_{dim}$ and the number of parameters $p_{no}$. After fixing the number of  GNN layers ($l_{no} = 4$), three different methods are employed. {\em Base}, the first one, fixes the hidden dimension $h_{dim} = 1000$ for all input features size $in_{dim}$. By doing so, the network size changes with the different sizes of the input elements.
In the method called {\em Padding}, instead, the size of the network is fixed by padding the input size and forcing it to have dimension: $in_{dim} = ( bbox + repr + max(Spacy, SciBERT)) = 861$. Finally {\em Scaled} fixes the number of network parameters $p_{no} = 100k$ by reshaping the hidden size $h_{dim}$ on the basis of the input size, according to the following second degree equation:
$p_{no} = (l_{no}-2)*h_{dim}^2 + (in_{dim}+out_{dim})*h_{dim}.$
The main difference between {\em Padding} and  {\em Scaled} is related to the first layer parameters which are not completely used by {\em Padding} since some input values are always set to zero.

\subsection{Results}
\label{results}
Experimental results (summarized in Table \ref{tab:experiment}) are designed to answer two main questions. 
Firstly, whether the proposed representation embedding improves the performance on table detection and discrimination of cells and headers.
Secondly, if the representation embedding is a good alternative to language models.
Concerning the first question, we can notice that in general by adding the representation embedding we have better performance for the detection of table cells (cell F1).
Model $B$ adds representation embeddings to the positional information of model $A$ and, in this case,
the cell F1 score increases for all methods (e.g. for Padding by 5.4 \%). 
cell-h F1 scores are lower because headers often contain words that are not well modeled by representation embeddings. 
Furthermore, if we add the SciBERT word embeddings to representations, we notice that model $F$ outperforms all the others in the detection of table cells (using the {\em Base} method). With respect to model $E$ the cell F1 score increases by 1.7\%.
Regarding the second question,  we can notice that representation embedding alone achieves good results.
For cell F1 score, model $B$ obtains almost the same results compared to $C$ (where Spacy replaces the representation embedding) and $E$ (where SciBERT replaces the representation embedding).
About the accuracy, model $B$ outperforms $C$  over all the methods and $E$ with padding.

The results can also be verified qualitatively by looking at inferences of two pages in Fig. \ref{fig:inference}. For each page, we present the graph node predictions (pages with tokens) and the grouped entities (pages with object boundaries). The majority of table headers and cell nodes are well recognized. Some errors are present mainly near region boundaries: e.g., at the top of table bodies, some table cells are misclassified as headers.
As a first step towards page objects identification, we apply a post-processing phase based on PyMuPDF \textit{blocks}. The library detects groups of entities that are then labeled with the majority class among them. Even if this approach works most of the times, relying on simple rules is prone to errors. In the rightmost image, it is possible to see that PyMuPDF fails because there is too much space withing the table and it outputs overlapping boxes.

\section{Conclusions}
\label{sec-concl}

We presented a Graph Neural Network model to perform document layout analysis and table extraction in scientific papers.
We propose a pipeline to represent PDF papers as graphs and redefine the problem of table extraction as a node classification problem by means of GNNs. 
We enhance node features using novel representation embeddings that we empirically prove to be effective to discriminate table elements from other classes. 
To carry out the experiments two widely-used datasets are merged.

Even if the post-processing provides promising results, to improve them future works will investigate and include edge classification in the GNN model to group elements belonging to same entities. In this way it  will be possible to compare this approach with both TD and TSR methods.
We also aim to further investigate the representation embedding by studying its properties.
 


\newpage

\balance

\bibliographystyle{IEEEtran}
\bibliography{
    IEEEabrv, 
    GeVi_ICPR22.bib
}



\end{document}